\documentclass[lettersize,journal]{IEEEtran}

\ifCLASSINFOpdf
\else
   \usepackage[dvips]{graphicx}
\fi

\usepackage{overpic}
\usepackage{url}
\usepackage{amsmath,amssymb,stmaryrd}   
\usepackage{algorithmic}
\usepackage{algorithm}
\usepackage{array}
\usepackage[caption=false,font=normalsize,labelfont=sf,textfont=sf]{subfig}
\usepackage{textcomp}
\usepackage{stfloats}
\usepackage{url}
\usepackage{verbatim}
\usepackage{graphicx}
\usepackage{cite}
\usepackage{booktabs}
\usepackage{hyperref}
\usepackage{pifont}
\usepackage{color}

\usepackage[capitalize]{cleveref}
\crefname{section}{Sec.}{Secs.}
\Crefname{section}{Section}{Sections}
\Crefname{table}{Table}{Tables}
\crefname{table}{Tab.}{Tabs.}

\usepackage{graphicx}

\begin{document}

\title{CM-UNet: Hybrid CNN-Mamba UNet for Remote Sensing Image Semantic Segmentation}
\author{
    Mushui~Liu,
    Jun~Dan,
    Ziqian~Lu,
    Yunlong~Yu*,
    Yingming Li, 
    Xi Li 
\thanks{This work is supported in part by the National Natural Science Foundation of China under Grant (62002320, U19B2043, 61672456), the Key R\&D  Program of Zhejiang Province, China (2021C01119).}
\thanks{Y. Yu* is the corresponding author.}
\thanks{M. Liu, J. Dan, and Y. Yu are with are with the College of Information Science and Electronic Engineering, Zhejiang University, Hangzhou, 310027 China (e-mail: \{lms,danjun,yuyunlong\}@zju.edu.cn).}
\thanks{Ziqian Lu is with the School of Aeronautics and Astronautics, Zhejiang University, Hangzhou 310027, China (e-mail: ziqianlu@zju.edu.cn)}
\thanks{Xi Li is with the College of Computer Science, Zhejiang University, Hangzhou 310027, China (e-mail: xilizju@zju.edu.cn)}
}


\markboth{Journal of \LaTeX\ Class Files, Vol. 14, No. 8, August 2015}
{Shell \MakeLowercase{\textit{et al.}}: Bare Demo of IEEEtran.cls for IEEE Journals}
\maketitle

\begin{abstract}
Due to the large-scale image size and object variations, current CNN-based and Transformer-based approaches for remote sensing image semantic segmentation are suboptimal for capturing the long-range dependency or limited to the complex computational complexity. In this paper, we propose CM-UNet, comprising a CNN-based encoder for extracting local image features and a Mamba-based decoder for aggregating and integrating global information, facilitating efficient semantic segmentation of remote sensing images. Specifically, a CSMamba block is introduced to build the core segmentation decoder, which employs channel and spatial attention as the gate activation condition of the vanilla Mamba to enhance the feature interaction and global-local information fusion. Moreover, to further refine the output features from the CNN encoder, a Multi-Scale Attention Aggregation (MSAA) module is employed to merge the different scale features. By integrating the CSMamba block and MSAA module, CM-UNet effectively captures the long-range dependencies and multi-scale global contextual information of large-scale remote-sensing images. Experimental results obtained on three benchmarks indicate that the proposed CM-UNet outperforms existing methods in various performance metrics. The codes are available at \href{https://github.com/XiaoBuL/CM-UNet}{https://github.com/XiaoBuL/CM-UNet}.
\end{abstract}

\begin{IEEEkeywords}
UNet, Visual State Space Model, Remote Sensing Semantic Segmentation
\end{IEEEkeywords}
\IEEEpeerreviewmaketitle

\section{Introduction} \label{sec:intro}
\IEEEPARstart{R}{emote} sensing image semantic segmentation involves classifying the pixels within the large-scale remote sense image into distinct categories to enhance the analysis and interpretation of remote sensing (RS) data. Such large-scale semantic segmentation is crucial for applications in autonomous driving \cite{chen2017multi}, urban planning \cite{liu2018semantic}, environmental protection \cite{chen2020sar}, and many other practical applications.

\begin{figure}
    \centering
    \begin{overpic}[width=\linewidth]{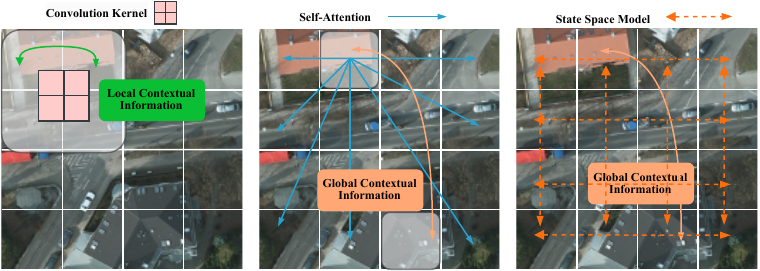}
        \put(8,-5){\small (a) CNN}
        \put(38,-5){\small (b) Transformer}
        \put(76,-5){\small (c) SSM}
    \end{overpic}
    \vspace{-0.01mm}
    \caption{Illustration of different architectures for processing remote images}
    \label{fig:intro1}
\end{figure}

With the advent of deep learning, the UNet \cite{ronneberger2015u}, has become the fundamental backbone network for segmentation tasks. UNet is well-known for its symmetrical U-shaped encoder-decoder architecture and integral skip connections, which effectively preserve key spatial information and merge features from the encoder and decoder layers for addressing the segmentation issues of complex structures. However, in the area of RS, where images often encompass large-scale scenes with significant target variations, UNet architectures built upon either convolutional neural networks (CNNs) \cite{diakogiannis2020resunet} or Transformers \cite{he2022swin} encounter limitations. They may struggle to capture global context adequately or exhibit high computational complexity, as illustrated in \cref{fig:intro1}~(a), (b). Therefore, it is essential to develop more efficient architectures capable of capturing comprehensive local-global information.

Recent developments have introduced the innovative architecture Mamba \cite{gu2023mamba}, which excels at efficiently capturing global contextual information. Mamba is designed for long-range modeling and is renowned for its computational efficiency based on state space models (SSM) \cite{gu2021efficiently}. Subsequently, Vision Mamba \cite{zhu2024vision} and VMamba \cite{liu2024vmamba} extend the architecture of Mamba to the field of computer vision, enhancing Mamba's unidirectional scanning mechanism. Considering the efficient global contextual modeling ability, the Mamba architecture is well-suitable for RS image processing, as shown in \cref{fig:intro1}~(c). PanMamba \cite{he2024pan}, RMamba \cite{chen2024rsmamba}, RS-Mamba \cite{zhao2024rs}, and RS3Mamba \cite{ma2024rs3mamba} have explored the application of Mamba for processing RS images. These methods either replace the network with Vision Mamba blocks and train from scratch or directly apply the pre-trained Vision Mamba blocks. However, they rarely consider the integration of local and global information within RS images, potentially limiting their ability to exploit the features provided by the well-pretrained CNN models fully.


In this paper, we propose CM-UNet, a novel framework for RS image semantic segmentation. CM-UNet leverages the Mamba architecture to aggregate multi-scale information from the CNN encoder. It comprises a U-shaped network with a CNN encoder extracting multi-scale textual information and a decoder featuring the designed CSMamba block for efficient semantic information aggregation. The CSMamba block utilizes the Mamba block for capturing long-range dependencies with linear time complexity and employs channel and spatial attention for feature selection. Serving as an alternative to previous self-attention transformer blocks, the CSMamba block enhances efficiency in RS semantic segmentation. Furthermore, a multi-scale attention aggregation (MSAA) module is introduced to ensemble features from different levels of the CNN encoder, aiding the CSMamba decoder through skip connections. Finally, CM-UNet incorporates multi-output supervision at various decoder levels to progressively generate semantic segmentation for RS images. The contributions are summarized as follows:

\begin{enumerate}
    \item We propose a mamba-based framework named CM-UNet to efficiently integrate the local-global information for RS image semantic segmentation. 
    \item We design a CSMamba block that encompasses the channel and spatial attention information into the mamba block to extract the global contextual information. Furthermore, we employ a multi-scale attention aggregation module to assist the skip connection and a multi-output loss to gradually supervise the semantic segmentation. 
    \item Extensive experiments on three well-known, publicly available RS datasets, i.e., ISPRS Potsdam, ISPRS Vaihingen, and LoveDA have shown the superiority of the proposed CM-UNet.
\end{enumerate}

\section{METHODOLOGY} \label{sec:method}

\begin{figure*}[!htb]
    \centering
    \includegraphics[width=\linewidth]{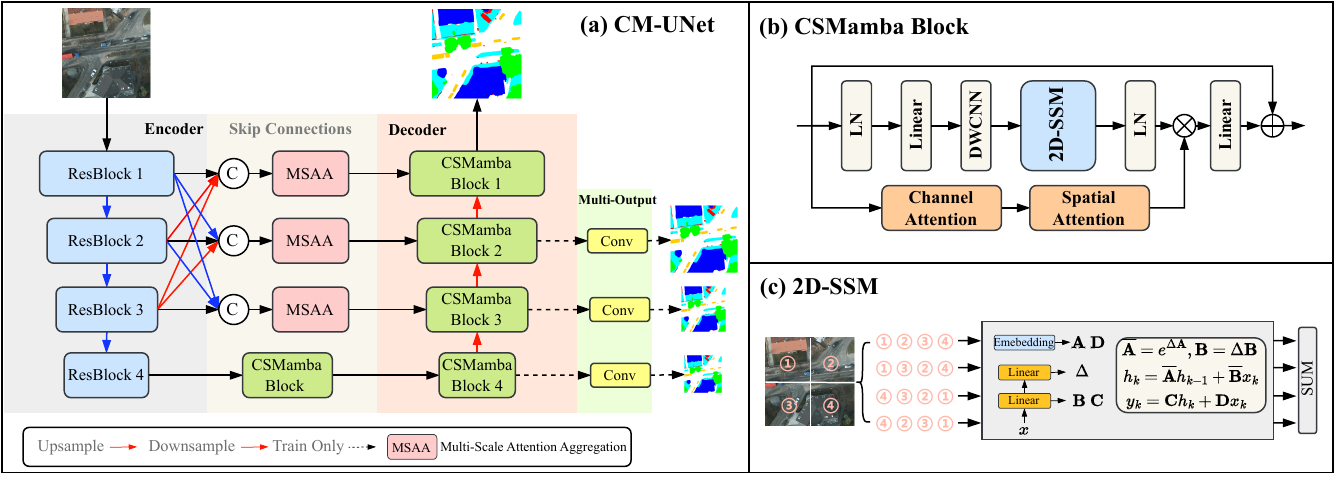}
    \caption{Framework of the proposed CM-UNet for remote sensing image semantic segmentation. (a) CM-UNet. (b) CSMamba Block. (c) 2D-SSM Module. }
    \label{fig:framework}
\end{figure*}

Our CM-UNet framework, illustrated in \cref{fig:framework}~(a), comprises three core components: a CNN-based encoder, the MSAA module, and the CSMamba-based decoder. The encoder employs ResNet to extract multi-level features, while the MSAA module fuses these features, replacing UNet's vanilla skip connections and enhancing the decoder's capability. In the CSMamba decoder, the assembly of CSMamba blocks aggregates local textual features to establish a comprehensive semantic understanding.

\subsection{CSMamba Block}

\begin{figure}
    \centering
    \includegraphics[width=\linewidth]{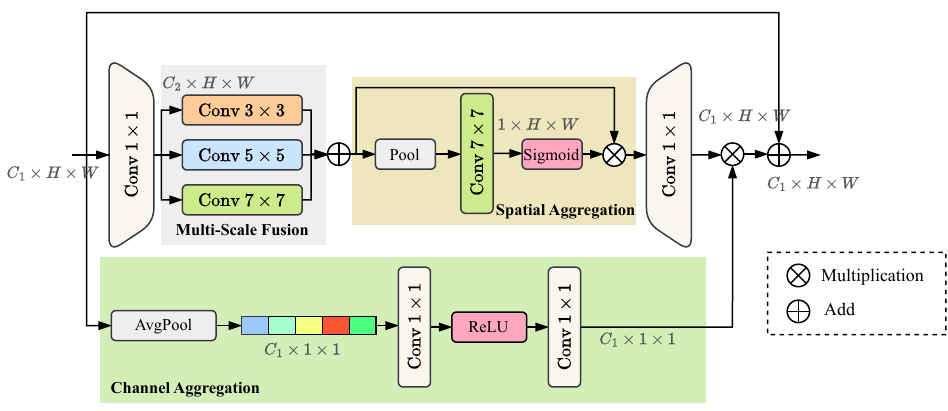}
    \caption{The Multi-Scale Attention Aggregation (MSAA) module.}
    \label{fig:msa}
\end{figure}

Large-scale RS semantic segmentation benefits from models that integrate both global and local information processing capabilities \cite{wang2022unetformer, wu2023cmtfnet}. Recent advancements, including transformer architectures \cite{dosovitskiy2020image,liu2021swin} utilizing self-attention mechanisms, have demonstrated significant efficacy in various visual tasks. However, these models often encounter a limitation in the form of square time complexity, which poses challenges for scalability and efficiency in processing large-scale RS images. This underscores the need for innovative approaches that maintain the strengths of transformers while mitigating their computational demands.

Motivated by the success of Mamba in long-range modeling with linear complexity, we introduce the Vision State-Space Module to the RS semantic segmentation area. Following~\cite{liu2024vmamba}, the input feature $X \in \mathbb{R}^{H \times W \times C}$ will go through two parallel branches. In the first branch, the feature channel is expanded to $\lambda C$ by a linear layer, where $\lambda$ is a pre-defined channel expansion factor, followed by a depth-wise convolution, SiLU activation function, together with the 2D-SSM layer and Layernorm. In the second branch, the features are integrated by the channel and spatial attention (CS) followed by the SiLU activation function. After that, features from the two branches are aggregated with the Hadamard product. Finally, the channel number is projected back to $C$ to generate output $X_{out}$ with the same shape as input:
\begin{equation} \label{eq:vssm}
\centering
\begin{aligned}
    &X_1 = \mathrm{LN(2D\text{-}SSM(SiLU(DWConv(Linear}(X)))))\\
    &X_2 = \mathrm{SiLU(CS}(X)),\\
    &X_{out} = \mathrm{Linear}(X_1 \odot X_2),
\end{aligned}
\end{equation}
where DWConv represents depth-wise convolution, CS means the channel and spatial attention module, 2D-SSM is the 2D selective scan module, and $\odot$ denotes the Hadamard product. The original Mamba model \cite{gu2023mamba} processes 1-D data via sequential selective scanning, which suits NLP tasks but challenges non-causal data forms like images. Following \cite{liu2024vmamba}, we incorporate the 2D Selective Scan Module (2D-SSM) for image semantic segmentation. As shown in \cref{fig:framework}.(c), 2D-SSM flattens the image feature into a 1D sequence and scans it in four directions: top-left to bottom-right, bottom-right to top-left, top-right to bottom-left, and bottom-left to top-right. This approach captures long-range dependencies in each direction via the selective state-space model. The direction sequences are then merged to recover the 2D structure. 

\subsection{Multi-Scale Attention Aggregation.}

\cref{fig:msa} depicts the Multi-Scale Attention Aggregation (MSAA) module for refining features in RS images. Outputs from the ResNet encoder stages, ${F}_{1}$, ${F}_{2}$, and ${F}_{3}$, are concatenated as $\hat{F}_{i} = \text{Concat}({F}_{i}, {F}_{i-1}, {F}_{i+1})$. The combined features $\hat{{F}} \in \mathcal{R}^{C_1 \times H \times W}$ are fed into the MSAA for refinement. Within the MSAA, dual paths—spatial and channel—are used for feature aggregation. Spatial refinement starts with channel projection via a $1 \times 1$ convolution reducing channel $C_1$ to $C_2$ where $C_2 = \frac{C_1}{\alpha}$. Multi-scale fusion involves summing convolutions across different kernel sizes, e.g., $3 \times 3, 5 \times 5, 7 \times 7$. Subsequently, spatial features are aggregated using mean and max pooling, followed by a $7\times7$ convolution and element-wise multiplication with the sigmoid-activated feature map.

In parallel, channel aggregation uses global average pooling to reduce dimensions to \( C_1 \times 1 \times 1 \), followed by $1 \times 1$ convolutions and ReLU activation to generate a channel attention map. This map is expanded to match the input's dimensions and combined with the spatially refined map. The MSAA thus enhances spatial and channel-wise features for subsequent network layers. By incorporating the MSAA module, the resulting feature maps are enriched with refined spatial and channel-wise information.

\subsection{Multi-Output Supervision.}
To effectively supervise the decoder in progressively generating the semantic segmentation map with RS images, our CM-UNet architecture incorporates intermediate supervision at each CSMamba block. This ensures that each stage of the network contributes to the final segmentation result, promoting more refined and accurate outputs. For the intermediate output of $i_{th}$ CSMamba block is
\begin{equation} \label{eq:multi-output}
p^{i} = \text{Conv}(F_{cs}^i),
\end{equation}
where $F_{cs}$ is the feature of $i_{th}$ CSMamba block. The $\text{Conv}$ module is used for mapping the features to the output $C$ channels class prediction map. Overall, the network is trained using a combination of standard cross-entropy loss and Dice loss.



\section{Experiments}\label{sec:exp}

\subsection{Datasets}
The proposed methodology is evaluated using the ISPRS Potsdam, ISPRS Vaihingen, and LoveDA \cite{wang2021loveda} RS segmentation datasets. For the ISPRS Potsdam dataset, 14 images are allocated for testing, while the remaining 23 images (excluding image $7_{10}$ due to erroneous annotations) are designated for training purposes. The ISPRS Vaihingen dataset comprises 12 patches for training and 4 patches for testing. Regarding the LoveDA dataset, the training set consists of 1,156 images, complemented by a test set of 677 images. We employ the mean F1 score (mF1), mean Intersection over Union (mIoU), and overall accuracy (OA) as the evaluation metrics.

\subsection{Implementation details}
We conduct all experiments on a single NVIDIA 3090 GPU with the PyTorch framework. We employ the AdamW optimizer with a base learning rate 6e-4 and the cosine strategy is adopted to adjust the learning rate. Following \cite{wang2022unetformer}, for the Vaihinge, Potsdam, and LoveDA datasets, the images were randomly cropped into $512 \times 512$ patches. For training, the augmentation techniques like random scale ($[0.5, 0.75, 1.0, 1.25, 1.5]$), random vertical flip, random horizontal flip and random rotate were adopted during the training process, while the training epoch was set as 100 and the batch size was 16. For the test stage, the test-time augmentation (TTA), e.g. vertical flip and horizontal flip are used. 

\subsection{Performance Comparison}
For the comparative analysis, we incorporate a selection of notable competitors as benchmarks, encompassing DeepLabV3+ \cite{chen2018encoder}, DANet \cite{fu2019dual}, ABCNet \cite{li2021abcnet}, BANet \cite{banet}, CMTFNet \cite{wu2023cmtfnet}, UNetformer \cite{wang2022unetformer}, ESDINet \cite{ESDINet}, BANet \cite{banet}, and Segmenter \cite{strudel2021segmenter}. These approaches utilize well-established encoder architectures, such as R18 \cite{he2016deep}, VMamba \cite{liu2024vmamba}, and Swin-Base \cite{liu2021swin}. 

\subsubsection{The ISPRS Potsdam dataset}
\begin{figure}[!ht]
    \centering
    \includegraphics[width=\linewidth]{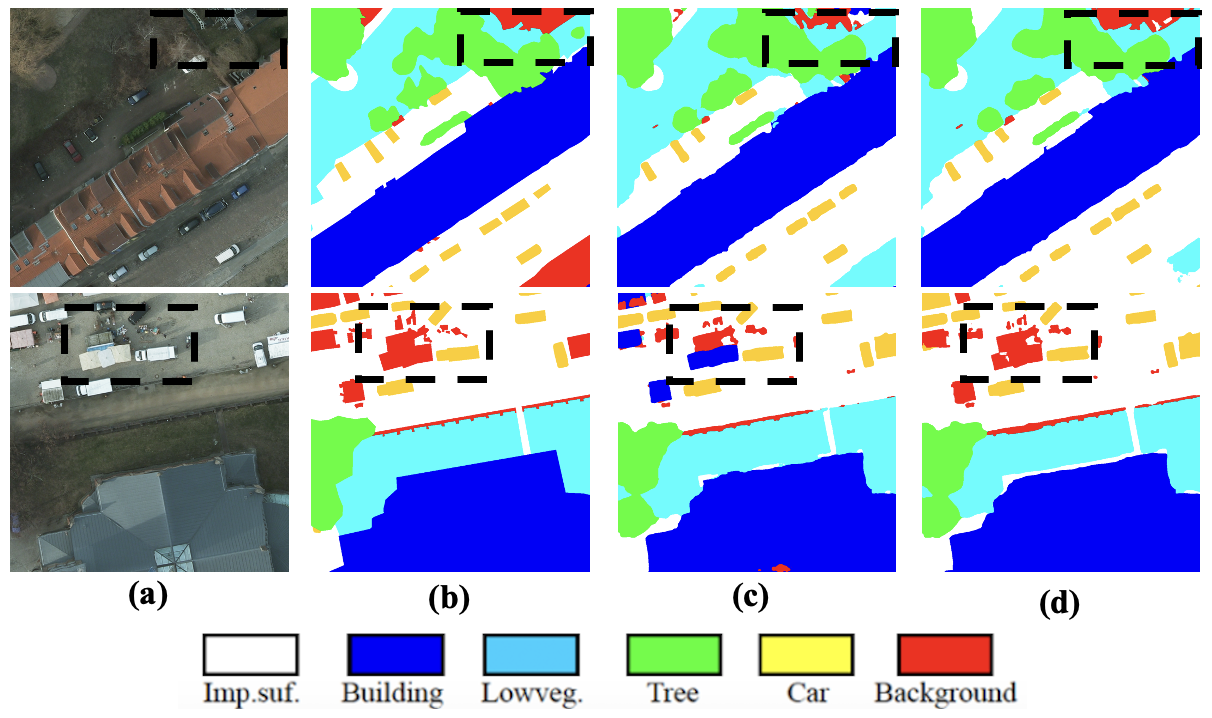}
    \vspace{-0.7cm}
    \caption{Visualization results on the ISPRS Postdam dataset. (a) NIRRG images. (b) GT. (c) UNetFormer. (d) Ours.}
    \label{fig:postdam}
\end{figure}

As shown in \cref{tab:potsdam}, CM-UNet outperforms competitors on the ISPRS Potsdam test set. Achieving an mF1 of 93.05\%, OA of 91.86\%, and mIoU of 87.21\%, it surpasses UNetformer by margins of 0.25\%, 0.56\%, and 0.41\%. Notably, it shows significant improvements over traditional methods like DANet and Segmenter, with 6.91\% and 6.51\% boosts in mIoU metric. These results highlight the effectiveness of its pre-trained ResNet backbone and innovative architecture for spatial feature learning. CM-UNet's superiority extends to various metrics compared to recent models like ESDINet, UNetformer, and CMTFNet, showcasing its versatility and efficacy. Qualitative comparisons illustrated in \cref{fig:postdam} further demonstrate its superiority over UNetformer, particularly in extracting sharper building outlines and reducing false segmentation.

\begin{table}[!ht]
    \centering
    \caption{Experimental results (\%) on the ISPRS Potsdam test set. Bold values are the best.}
    \resizebox{\linewidth}{!}{
        \begin{tabular}{l|c|ccccc|ccc}
        \toprule
        Method & Backbone & Imp. surf. & Building & Low. veg. & Tree & Car & mF1 & OA & mIoU \\
        \midrule
        DANet \cite{fu2019dual} & R18 & 91.00 & 95.60 & 86.10 & 87.60 & 84.30 & 88.90 & 89.10 & 80.30 \\
        ABCNet  \cite{li2021abcnet} & R18 & 93.50 & 96.90 & 87.90 & 89.10 & 95.80 & 92.70 & 91.30 & 86.50 \\
        BANet  \cite{banet} & ResT & 93.34 & 96.66 & 87.37 & 89.12 & 95.99 & 92.50 & 91.06 & 86.25 \\ 
        Segmenter \cite{strudel2021segmenter} & ViT-T & 91.50 & 95.30 & 85.40 & 85.00 & 88.50 & 89.20 & 88.70 & 80.70 \\
        ESDINet \cite{ESDINet} & R18 & 92.68 & 96.28 & 87.29 & 88.10 & 95.42 & 91.96 & 90.53 & 85.32 \\
        UNetformer \cite{wang2022unetformer} & R18  & 93.60 & \textbf{97.20} & 87.70 & 88.90 & \textbf{96.50} & 92.80 & 91.30 & 86.80 \\
        CMTFNet \cite{wu2023cmtfnet} & R50 & 92.12 & 96.41 & 86.43 & 87.26 & 92.41 & 90.93 & 89.89 & 83.57 \\
        \midrule
        CM-UNet  & R18 & \textbf{94.34} & 96.98 & \textbf{88.18} & \textbf{89.41} & 96.34 & \textbf{93.05} & \textbf{91.86} & \textbf{87.21} \\
        \bottomrule
        \end{tabular}
    }
    \label{tab:potsdam}
\end{table}

\subsubsection{The ISPRS Vaihingen Dataset}
\begin{figure}[!t]
    \centering
    \includegraphics[width=\linewidth]{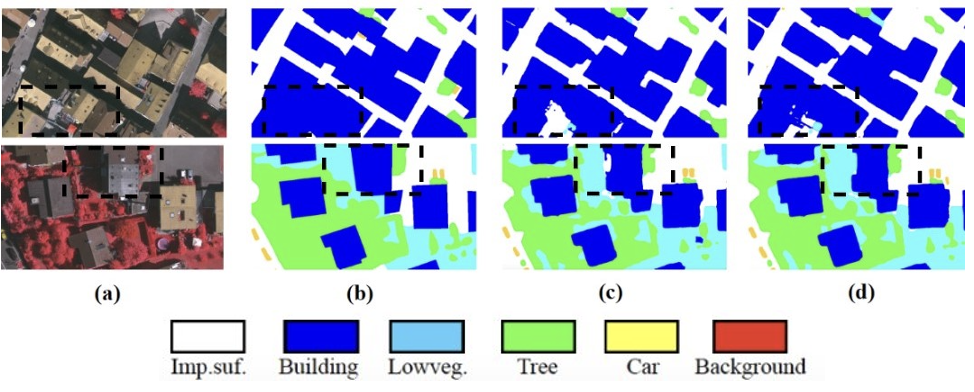}
    \vspace{-0.7cm}
    \caption{Visualization results on the ISPRS Vaihingen dataset. (a) NIRRG images. (b) GT. (c) UNetFormer. (d) Ours.}
    \label{fig:vaihingen}
\end{figure}

\cref{tab:vaihingen} showcase the experimental results. CM-UNet achieves mIoU of 85.48\%, surpassing competitors by 2.78\% to 16.08\%. Its mF1 (92.01\%) and OA (93.81\%) also compare favorably. In F1 scores, CM-UNet excels in multiple categories, notably Imp.surf., Building, Low.veg., and Car. It outperforms UNetFormer by 4.42\% in Imp.surf. This underscores Mamba's ability to capture irregular objects and global-local relationships crucial in RS. Visualization as shown in \cref{fig:vaihingen} confirms its accuracy, especially in discerning anomalous samples and subtle changes like shadows. CM-UNet's precise predictions, particularly in imp. surf. and building categories, underscore its ability to perceive global long-range facts and spatial contextual features.

\begin{table}[!t]
    \centering
    \caption{
    Experimental results (\%) on the ISPRS Vaihingen Urban dataset. Bold values are the best.}
    \resizebox{\linewidth}{!}{
        \begin{tabular}{l|c|ccccc|ccc}
        \toprule
        Method & Backbone & Imp. surf. & Building & Low.veg. & Tree & Car & mF1 & OA & mIoU \\
        \midrule
        DANet \cite{fu2019dual} & R18 & 90.00 & 93.90 & 82.20 & 87.30 & 44.50 & 79.60 & 88.20 & 69.40 \\
        ABCNet  \cite{li2021abcnet} & R18 & 92.70 & 95.20 & 84.50 & 89.70 & 85.30 & 89.50 & 90.70 & 81.30 \\
        BANet  \cite{banet} & ResT & 92.23 & 95.23 & 83.75 & 89.92 & 86.76 & 89.58 & 90.48 & 81.35 \\ 
        Segmenter \cite{strudel2021segmenter} &  ViT-T & 89.80 & 93.00 & 81.20 & 88.90 & 67.60 & 84.10 & 88.10 & 73.60 \\ 
        ESDINet \cite{ESDINet} & R18 & 92.74 & 95.54 & 84.46 & 90.03 & 87.16 & 89.99 & 90.88 & 82.03 \\
        UNetformer \cite{wang2022unetformer} & R18  & 
        92.70 & 95.30 & 84.90 & 90.60 & 88.50 & 90.40 & 91.00 & 82.70 \\
        FTUNetformer \cite{wang2022unetformer} & Swin-B  & 93.50 & 96.00 & 85.60 & \textbf{90.80} & 90.40 & 91.30 & 91.60 & 84.10 \\
        CMTFNet \cite{wu2023cmtfnet} & R50  & 90.61 & 94.21 & 81.93 & 87.56 & 82.77 & 87.42 & 88.71 & 77.95 \\ 
        RS3Mamba \cite{ma2024rs3mamba} & R18-Mamba & 96.70 & 95.50 & 84.40 & 90.00 & 86.90 & 90.70 & 93.20 & 83.30 \\
        \midrule
        CM-UNet  & R18 & \textbf{97.12} & \textbf{96.20} & \textbf{85.65} & 90.51 & \textbf{90.58} & \textbf{92.01} & \textbf{93.81} & \textbf{85.48} \\
        \bottomrule
        \end{tabular}
    }
    \label{tab:vaihingen}
\end{table}

\subsubsection{The LoveDA Dataset}
\begin{figure}[!t]
    \centering
    \includegraphics[width=\linewidth]{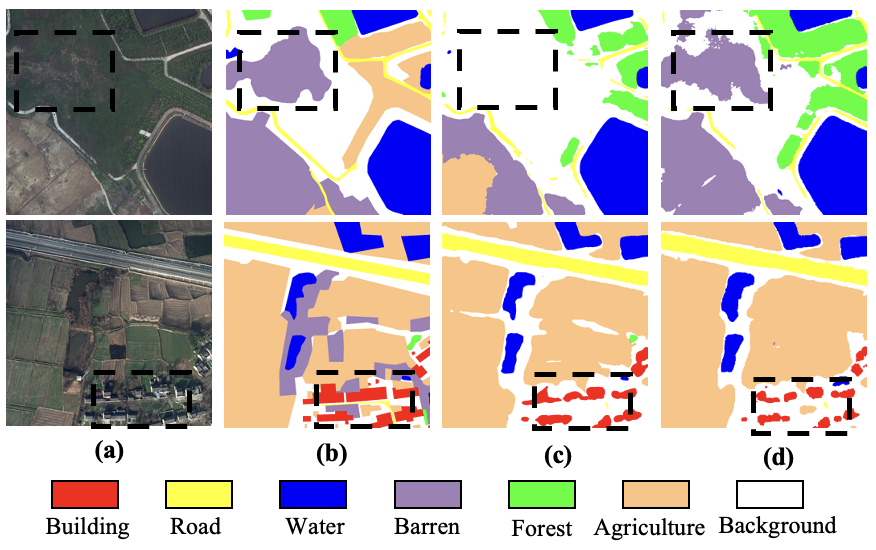}
    \vspace{-0.7cm}
    \caption{Visualization results on the LoveDA dataset. (a) NIRRG images. (b) GT. (c) UNetFormer. (d) Ours.}
    \label{fig:loveda}
\end{figure}

\cref{tab:loveda} shows the results on the LoveDA dataset. Notably, Our approach achieves performance, with mIoU 52.17\%. Besides, CM-UNet excels in various classes, e.g.,  background, building, and road. Visualization shown in \cref{fig:loveda} highlights CM-UNet's superiority in delineating land cover classes compared to UNetFormer. It accurately captures building edges, roadways, and agricultural areas, approaching ground truth labels with precision even in complex urban scenes and intricate agricultural patterns. The method's consistency across different classes underscores its higher classification accuracy and enhanced edge detection capability, crucial for precise land cover mapping. These results affirm the effectiveness of CM-UNet on large-scale RS images.

\begin{table}[!t]
  \centering
	\caption{Experimental results (\%) on the LoveDA Urban dataset. Bold values are the best.}
	  \tabcolsep=0.12cm
	  \renewcommand\arraystretch{1.5}
   \resizebox{\linewidth}{!}{
        \begin{tabular}{l|c|ccccccc|c}
        \toprule
       {Method}  & Backbone & {Background} & {Building} & {Road}  & {Water}  & {Barren}  & {Forest} & {Agriculture} & {mIoU}  \\
        \midrule
        DeepLabV3+ \cite{chen2018encoder} & R50 & 43.00& 50.90& 52.00& 74.40& 10.40& 44.20& 58.50& 47.60 \\
        Segmenter \cite{strudel2021segmenter} &  ViT-T & 38.00& 50.70& 48.70& 77.40& 13.30& 43.50& 58.20& 47.10 \\
        ABCNet \cite{li2021abcnet} & R50 & 53.00 & 62.18 & 52.42 & 62.02 & 29.80 & 41.92 & 47.27 & 49.80\\
        BANet \cite{bo2022basnet} & Res-T & 53.94 & 62.14 & 51.33 & 64.59 & 27.07 & 43.86 & 48.12 & 50.15\\
        UNetformer  \cite{wang2022unetformer} & R18 & 44.70 & 58.80& 54.90& 79.60& 20.10& 46.00& 62.50 & 50.73 \\
        \midrule
        CM-UNet  & R18 & \textbf{54.56} & \textbf{64.13} & \textbf{55.51} & 68.06 & 29.62 & 42.92 & 50.42 & \textbf{52.17} \\
         \bottomrule
	\end{tabular}
    }
    \label{tab:loveda}
\end{table}

\subsection{Further Analysis}

\subsubsection{Effect of Model Architecture} The ablation study on the designed blocks is as depicted in \cref{tab:smsa}. Specifically, incorporating the Multi-Scale Attention Aggregation (MSAA) module alone leads to improvements, indicating its efficacy in capturing contextual information across different scales. Similarly, integrating the Multi-output strategy further enhances segmentation performance, demonstrating the benefits of leveraging multiple prediction outputs. Notably, the combined use of both MSAA and Multi-output modules yields the highest gains in all metrics, highlighting the synergistic effects of these architectural components in enhancing the model's ability to discern intricate spatial features and improve overall segmentation accuracy.

\begin{table}[!t]
    \centering
    \caption{Ablation results (\%) on the designed blocks.}
    \resizebox{0.8\linewidth}{!}{
    \begin{tabular}{cc|ccc}
    \toprule
    MSAA & Multi-output & mF1 & OA & mIoU\\ 
    \midrule
   \ding{56} &\ding{56} & 91.31 & 90.78 & 83.53\\ 
    \ding{52}  & \ding{56} & 92.41 & 91.23  & 84.45 \\
   \ding{56} & \ding{52} & 91.89 & 90.89 & 83.91 \\  
    \ding{52} & \ding{52} & 93.05 & 91.86 & 85.48 \\
    \bottomrule
    \end{tabular}
    }
    \label{tab:smsa}
\end{table}

\begin{table}[!t]
    \centering
    \caption{Computational complexity analysis measured by two $256 \times 256$ images on a single NVIDIA 3090 GPU. MIoU value is evaluated on the ISPRS VAIHINGEN DATASET.}
    \resizebox{\linewidth}{!}{
    \begin{tabular}{l|cccc}
    \toprule
    {Model} & {FLOPs (G) $\downarrow$}  &  {Param. (M) $\downarrow$} & {Memory (MB) $\downarrow$} &  {mIoU(\%) $\uparrow$}  \\
    \midrule
   ABCNet \cite{li2021abcnet} & {7.81} & {13.39} & {306.18} & 81.30\\
    CMTFNet \cite{wu2023cmtfnet}   & {17.14} & {30.07} & {999.25} & 77.95 \\
    UNetformer \cite{wang2022unetformer} & {5.87} & {11.69} & {313.82} & 82.70 \\
    FTUNetformer \cite{wang2022unetformer}  & {50.84} & {96.14} & {1474.96} & 84.10 \\
    \midrule
    CM-UNet  & 6.01 & 12.89 & 366.33 & 85.48\\
        \bottomrule
    \end{tabular}
    }
    \label{tab:compute-efficiency}
\end{table}

\subsubsection{Model Complexity} \cref{tab:compute-efficiency} illustrates the comparison of model complexities using three metrics: floating-point operation count (FLOPs), model parameters, and memory footprint. Remarkably, CM-UNet achieves a harmonious balance across these metrics, exhibiting lower FLOPs and parameter counts alongside a relatively modest memory footprint. Despite this, its mIoU result surpasses that of other models, highlighting its favorable cost-effectiveness.

\section{Conclusion} \label{sec:conclusion}
In this paper, we introduce CM-UNet, an efficient framework that leverages the recent Mamba architecture for RS semantic segmentation. Our design addresses the significant target variations within large-scale RS images by featuring a novel UNet-shaped structure. The encoder utilizes ResNet to extract textual information, while the decoder employs CSMamba blocks to effectively capture global long-range dependencies. Additionally, we have integrated a Multi-Scale Attention Aggregation (MSAA) module and a multi-output enhancement to further support multi-scale feature learning. The CM-UNet has been validated across three RS semantic segmentation datasets, with experimental results demonstrating the superiority of our approach.

\bibliographystyle{IEEEtran}
\bibliography{references}

\end{document}